\documentclass[conference]{IEEEtran}
\IEEEoverridecommandlockouts
\usepackage{cite}
\usepackage{amsmath,amssymb,amsfonts}
\usepackage{algorithmic}  % or keep algorithm2e if you adapt alg commands
\usepackage{graphicx}
\usepackage{textcomp}
\usepackage{xcolor}
\usepackage{booktabs}
\usepackage{siunitx}
\usepackage{subcaption}
\usepackage{hyperref}
\def\BibTeX{{\rm B\kern-.05em{\sc i\kern-.025em b}\kern-.08em
		T\kern-.1667em\lower.7ex\hbox{E}\kern-.125emX}}
\usepackage{svg}

\usepackage[ruled,vlined]{algorithm2e}
\usepackage[utf8]{inputenc}
\DeclareUnicodeCharacter{202F}{\,}

\usepackage{cite}
\usepackage{amsmath,amssymb,amsfonts}
\usepackage{algorithm2e}
\usepackage{graphicx}
\usepackage{textcomp}
\usepackage{xcolor}
\usepackage{booktabs}
\usepackage{siunitx}
\usepackage{hyperref}
\usepackage{svg}
\usepackage{caption}
\usepackage{subcaption}
\usepackage{booktabs}
\usepackage{adjustbox}
\usepackage{dsfont}

\SetKwInput{KwIn}{Input}
\SetKwInput{KwOut}{Output}
\SetKwBlock{Pre}{\textbf{Preprocess:}}{}
\SetKwBlock{StageA}{\textbf{Stage 1: Clustering at $t=48$}}{}
\SetKwBlock{StageB}{\textbf{Stage 2: Clustering at $t=24$}}{}
\SetKwBlock{Post}{\textbf{Return}}{}

\usepackage{tikz}
\usetikzlibrary{positioning,arrows.meta,calc}

\title{Diffusion-Based Scenario Tree Generation for Multivariate Time Series Prediction and Multistage Stochastic Optimization
	
\thanks{This research was supported by the project “Applied Research for Autonomous Robotic Systems” (MIS5200632), funded by the European Union-NextGenerationEU under the National Recovery and Resilience Plan “Greece 2.0”.}
}

\author{Stelios Zarifis\textsuperscript{1,2}, Ioannis Kordonis\textsuperscript{3}, and Petros Maragos\textsuperscript{1,2,3}\\[0.5em]
	\textsuperscript{1}HERON - Hellenic Robotics Center of Excellence, Athens, Greece \\
	\textsuperscript{2}Robotics Institute, Athena Research Center, Athens, Greece \\
	\textsuperscript{3}School of Electrical \& Computer Engineering, National Technical University of Athens, Greece \\ [0.5em]
	\texttt{s.zarifis@athenarc.gr, kordonis@central.ntua.gr, petros.maragos@athenarc.gr}
}

\begin{document}

\maketitle

\begin{abstract}
	Stochastic forecasting is critical for efficient decision-making in uncertain systems, such as energy markets and finance, where estimating the full distribution of future scenarios is essential. We propose \emph{Diffusion Scenario Tree} (\emph{DST}), a general framework for constructing scenario trees using diffusion-based probabilistic forecasting models to provide a structured model of system evolution for control tasks. \emph{DST} recursively samples future trajectories and organizes them into a tree via clustering, ensuring non-anticipativity (decisions depending only on observed history) at each stage, offering a superior representation of uncertainty compared to using predictive models solely for forecasting system evolution. We integrate \emph{DST} into Model Predictive Control (MPC) and evaluate it on energy arbitrage in New York State’s day-ahead electricity market. Experimental results show that our approach significantly outperforms the same optimization algorithms that use scenario trees generated by more conventional models. Furthermore, using \emph{DST} for stochastic optimization yields more efficient decision policies by better handling uncertainty than deterministic and stochastic MPC variants using the same diffusion-based forecaster, and simple Model-Free Reinforcement Learning (RL) baselines.
\end{abstract}

\begin{IEEEkeywords}
	Diffusion Models, Scenario Trees, Stochastic Optimization, Energy Markets, Uncertainty Quantification
\end{IEEEkeywords}

% \vspace{-.5em}
\section{Introduction}
\label{sec:intro}
% \vspace{-.5em}
Robust decision-making under uncertainty requires forecasts that capture not only the most likely outcome, but also the full distribution of future evolution. Scenario trees, a classical tool in stochastic optimization, model time-evolving uncertainty while enforcing non-anticipativity, ensuring decisions at each node depend only on information revealed up to that point~\cite{Shapiro2021,Pflug2014}. Traditional methods like VAR and LSTM are widely used but struggle with complex, multimodal dynamics, typically offering point forecasts rather than distributions, which carry richer information for decision-making. In contrast, diffusion-based forecasters like TimeGrad~\cite{Rasul2021}, can effectively guide multistage stochastic optimization by learning the full conditional distribution of future observations. Other diffusion-based models for time series, such as those designed for financial denoising~\cite{WangVentre2024} or integrated with foundation models~\cite{LiuEtAl2024}, can similarly capture complex, temporal, multimodal dependencies. Diffusion-based methods have been applied to scenario generation in energy domains. For example, Li et al.~\cite{Li2025} use a multi‐resolution diffusion model with time-series decomposition for renewable energy outputs, and Zhao et al.~\cite{Zhao2025} propose a conditional diffusion model to jointly handle source-load uncertainties with spatio-temporal fusion, controllability and noise estimation. However, these works generate unstructured sets of scenarios rather than hierarchical scenario trees with explicit branching structure and non-anticipativity guarantees, and they do not demonstrate direct integration of such trees into multistage stochastic optimization or decision-making frameworks.

In this work, we introduce the \emph{Diffusion Scenario Tree (DST)} framework, a generic methodology for constructing scenario trees using a diffusion-based probabilistic forecaster. Each node represents a potential future state along with its associated realization probability, recursively generated from the learned distribution. This structure can be integrated naturally in optimization frameworks, providing the decision-maker with a rich representation of the forecast uncertainty.

We demonstrate the effectiveness of \emph{DST} by applying it to energy arbitrage, a multistage stochastic optimization problem in which a Battery Energy Storage System (BESS) participates in New York State's day-ahead electricity market. We compare the performance of \emph{DST}-guided stochastic MPC against scenario trees generated from VAR and LSTM models, our prior diffusion-based methods, and simple model-free Reinforcement Learning baselines.

Our previous work~\cite{Zarifis2025} introduced \emph{Diffusion-Informed MPC}, a framework that integrates diffusion forecasts directly into MPC algorithms. The present work extends this line of research by proposing a new algorithm that organizes diffusion-based predictions into non-anticipative \emph{scenario trees}. Our primary goal with \emph{DST} is to bridge the machine learning and stochastic optimization/operations research communities by offering a novel approach to structured uncertainty modeling. Evaluated on the same setup, \emph{DST} yields more efficient policies than standard MPC variants using the same diffusion forecaster, scenario-tree approaches based on conventional models and simple RL baselines, demonstrating the value of hierarchical scenario discretization in real-world control under partial observability.

% \vspace{-.5em}
\section{Background}
\label{sec:background}
% \vspace{-.5em}
Decision-making in applications like energy markets, finance, and robotics require precise modeling of time-evolving uncertainty to optimize actions. Scenario trees, a standard tool used in multistage stochastic optimization, represent uncertainty through nodes for possible future states and branches with associated probabilities~\cite{Shapiro2021,Pflug2014}. These trees respect non-anticipativity so that decisions at each stage depend only on observed information up to that point, and they discretize continuous distributions, enabling tractable optimization.

Early methods for scenario tree generation use sampling and scenario reduction~\cite{Shapiro2021,Rathi2023}, while more recent approaches rely on Markov chain approximations or moment-matching to capture multivariate dynamics~\cite{Pflug2014,Bertsimas2019}. Forecasting models are often used to generate scenarios across stages. Time series models like ARIMA or GARCH generate scenarios from historical data, followed by recursive reduction or bundling~\cite{Kaut2009,Heitsch2009}. Data-driven methods align scenarios with statistical properties such as mean and variance through optimization and clustering~\cite{Kaut2020}. Non-parametric density estimation, like kernel methods, samples trajectories without fixed forms, capturing complex patterns~\cite{Pichler2025}, while parametric models such as SDEs for price dynamics assume structured distributions~\cite{Kirui2020}. Beyond these, recent applications include learning-based adaptive scenario-tree MPC with Bayesian neural networks~\cite{Bao2023}, scenario-tree MPC-based RL for autonomous vehicles~\cite{BahariKordabad2021}, scenario-tree MPC for highway interactions~\cite{Alipour2024}, microgrid sizing with scenario trees~\cite{Bagheri2022}.

However, these methods face challenges: statistical forecasters assume normal distributions for the errors, limiting their ability to model complex patterns; traditional forecasting models do not have the required inherent complexity to capture multimodal dynamics; scenario reduction can compromise tree robustness by discarding useful scenarios; and RL-based approaches typically require vast amounts of data and offer no guarantee of convergence or stable training. Recent works have explored diffusion models in RL, e.g., for trajectory generation and planning in offline settings~\cite{janner2022planning}, as policies~\cite{wang2022diffusion}, for synthetic data augmentation~\cite{lu2023synthetic}.

In contrast, we propose to use diffusion-based forecasters, like TimeGrad~\cite{Rasul2021}, to acquire accurate, probabilistic predictions for the future of multivariate time series. TimeGrad, is an autoregressive model that guides the diffusion processes of a Denoising Diffusion Probabilistic Model (DDPM)~\cite{Ho2020} using a Recurrent Neural Network to encode the time series history, learning the conditional distribution of future multivariate time series:
\vspace{-1em}

\begin{equation}
	p_\theta(\mathbf{x}_{k_0:N} | \mathbf{x}_{1:k_0-1}) \approx q(\mathbf{x}_{k_0:N} | \mathbf{x}_{1:k_0-1}),
\end{equation}

% \vspace{-.5em}
where \(\mathbf{x}_k \in \mathbb{R}^D\) is the multivariate time series vector at time \(k\). TimeGrad has showcased impressive ability to capture long-range dependencies and multimodal patterns, making it ideal for sampling trajectories to construct scenario trees.

Our framework integrates TimeGrad’s probabilistic forecasts into scenario tree construction, using clustering to organize sampled trajectories while preserving non-anticipativity. This tree-based representation enables more robust multistage stochastic optimization compared to using point estimates or direct Monte Carlo sampling from the same forecaster. Empirical evidence from our energy arbitrage experiments with various time series forecasters shows that organizing predictions into scenario trees yields better policies than non-tree approaches using the exact same models. This pattern underscores the value of structured uncertainty discretization, particularly for forecasters that are less capable of capturing multimodal dynamics on their own.
% \vspace{-.5em}
\section{Methodology}
% \vspace{-.5em}
\subsection{Problem Definition}
% \vspace{-.5em}
We present a method for a class of partially observable stochastic systems, where dynamics are separable into a deterministic component, influenced by actions, and a stochastic one, driven by external factors. This class of problems with separable state is common across many real-world applications such as energy market trading, finance, inventory management, and robotics. We model the system as a Partially Observable Markov Decision Process (POMDP), defined by \((S, A, T, R, \Omega, O, \gamma)\), where \(S\) is the state space, \(A\) the action space, \(T\) the transition function, \(R\) the reward function, \(\Omega\) the observation space, \(O\) the observation function, and \(\gamma \in [0,1)\) the discount factor, consistent with our previous work~\cite{Zarifis2025}.

At time step \(k\), the environment is in a hidden state \(s_k \in \mathcal{S}\). The agent selects an action \(a_k \in \mathcal{A}\), causing the environment to transition to state \( s_{k+1} \), yielding observation \(o_{k+1} \sim O(o_{k+1} | s_{k+1}, a_k)\) and reward \(r_k = R(o_k, a_k)\). The observation is \(o_k = (o_k^d, o_k^s) \in \Omega\), where \(o_k^d\) is the deterministic part (e.g., battery state-of-charge) evolving as \(o_{k+1}^d = f(o_k^d, a_k)\), and \(o_k^s \in \mathbb{R}^D\) the stochastic part (e.g., energy prices). In applications like energy bidding, true states are unmeasurable due to unpredictable factors (e.g., demand, weather, human factor), so the agent must rely on observations.

The agent’s objective at time \(k_0\), given the observation history \(o_1, \dots, o_{k_0}\), is to find a policy \(\pi\) maximizing the expected cumulative discounted reward over a finite horizon:
\vspace{-.5em}
 
\begin{equation}
	\underset{\pi}{\text{maximize}} \quad \mathbb{E}\left[ \sum_{k=k_0}^{k_0+N-1} \gamma^{k-k_0} R(o_k^c, o_k^u, a_k) \right],
\end{equation}

% \vspace{-.5em}
which we propose to approximate with a finite-horizon multistage scenario tree-based MPC problem:
\vspace{-.5em}

\begin{equation}
	\underset{\pi}{\text{maximize}} \quad \left[ \sum_{k=k_0}^{k_0+N-1} \gamma^{k-k_0} R(o_k^d, o_k^s, a_k) \right],
\end{equation}

% \vspace{-.5em}
where, the agent forecasts the stochastic part \(\hat{o}_{k_0+1:k_0+N}^s\), organizing it into a scenario tree for tree-based optimization.
% \vspace{-.5em}
\subsection{Diffusion-Based Forecasting}
% \vspace{-.5em}
We use TimeGrad, a diffusion-based autoregressive model, to forecast the stochastic component’s conditional distribution \(p_\theta(o_{k+1}^s | o_{1:k}^s)\). At each decision epoch, TimeGrad generates \(M\) sample forecasts, and organizes them in tree nodes.
% \vspace{-.5em}
\subsection{Scenario Tree-Based Stochastic Optimization}
\label{sec:STMPC}
\begin{figure}[htbp]
\centering
\resizebox{.8\columnwidth}{!}{
	\begin{tikzpicture}[
		node distance = 1.8cm and 4cm,
		state/.style = {circle, draw, minimum size=0.8cm, fill=blue!10, font=\scriptsize},
		transition/.style = {draw, -{Stealth[length=3mm]}, thick},
		label/.style = {font=\small, align=center}
		]
		
		% Root node (current observation)
		\node[state, label=left:{\textbf{Stage $t$}}] (root) {$\mathbf{o}_{t}$};
		
		% First level (possible observations at stage t+1)
		\node[state, above right=1.2cm and 4cm of root] (o11) {$\mathbf{\hat{o}}_{t+1}^{(1)}$};
		\node[state, below right=1.2cm and 4cm of root] (o12) {$\mathbf{\hat{o}}_{t+1}^{(2)}$};
		
		% Second level (possible observations at stage t+2)
		\node[state, above right=.55cm and 4cm of o11] (o21) {$\mathbf{\hat{o}}_{t+2}^{(1)}$};
		\node[state, below right=.55cm and 4cm of o11] (o22) {$\mathbf{\hat{o}}_{t+2}^{(2)}$};
		\node[state, above right=.55cm and 4cm of o12] (o23) {$\mathbf{\hat{o}}_{t+2}^{(3)}$};
		\node[state, below right=.55cm and 4cm of o12] (o24) {$\mathbf{\hat{o}}_{t+2}^{(4)}$};
		
		% Root to first level transitions (stage t -> t+1)
		\draw[transition] (root) -- (o11)
		node[above, midway, label, sloped] { $P(\mathbf{\hat{o}}_{t+1}^{(1)}, w_{t+1}^{(1)})$ }
		node[below, midway, label, sloped] { $a_{t}$ };
		\draw[transition] (root) -- (o12)
		node[above, midway, label, sloped] { $P(\mathbf{\hat{o}}_{t+1}^{(2)}, w_{t+1}^{(2)})$ }
		node[below, midway, label, sloped] { $a_{t}$ };
		
		% First to second level transitions (stage t+1 -> t+2)
		\draw[transition] (o11) -- (o21)
		node[above, midway, label, sloped] { $P(\mathbf{\hat{o}}_{t+2}^{(1)}, w_{t+2}^{(1)})$ }
		node[below, midway, label, sloped] { $a_{t+1}^{(1)}$ };
		\draw[transition] (o11) -- (o22)
		node[above, midway, label, sloped] { $P(\mathbf{\hat{o}}_{t+2}^{(2)}, w_{t+2}^{(2)})$ }
		node[below, midway, label, sloped] { $a_{t+1}^{(1)}$ };
		\draw[transition] (o12) -- (o23)
		node[above, midway, label, sloped] { $P(\mathbf{\hat{o}}_{t+2}^{(3)}, w_{t+2}^{(3)})$ }
		node[below, midway, label, sloped] { $a_{t+1}^{(2)}$ };
		\draw[transition] (o12) -- (o24)
		node[above, midway, label, sloped] { $P(\mathbf{\hat{o}}_{t+2}^{(4)}, w_{t+2}^{(4)})$ }
		node[below, midway, label, sloped] { $a_{t+1}^{(2)}$ };
		
		% Labels for stages
		\node[below=.5cm of root, label, text width=3cm] {
			\begin{tabular}{c}
				Realized observation \\
				at stage $t$
			\end{tabular}
		};
		\node[below=.5cm of o12, label, text width=3cm] {
			\begin{tabular}{c}
				Realized observation \\
				at stage $t+1$
			\end{tabular}
		};
		% The node with the new position using shifts
		\node[anchor=south, yshift=-1.25cm, xshift=-1cm] at (o24) {Realized observation at stage $t+2$};
		
	\end{tikzpicture}
	}
	% \vspace{-1em}
	\caption{Visualization of a Scenario Tree that discretizes the probability distribution of the future steps.}
	\label{fig:scenario_tree_diagram}
	% \vspace{-1em}
\end{figure}
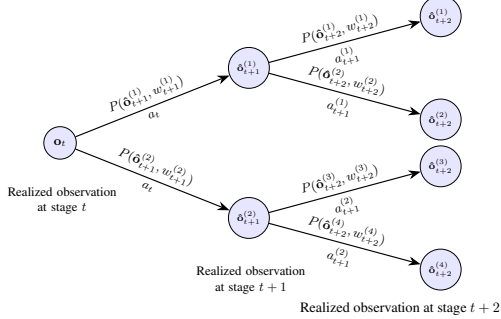

% \vspace{-.5em}
Scenario Tree-based MPC optimizes decisions over a finite horizon using a scenario tree, with the root as the current observation \(o_{k_0}\), nodes at each decision stage \(t\) representing possible observation realizations, and branches are associated with probabilities of transitioning to each node~\cite{rakovic2018handbook}. It is important to note that the branching frequency may differ from the observation time step frequency (e.g., in the experiments discussed in Section \ref{sec:results} the branching step is daily while the observations are hourly). Therefore, a node in the scenario tree may represent a sequence of observations, all acquired at one decision stage. Non-anticipativity ensures decisions at stage \(t\) depend solely on information available up to \(t\), enforcing identical actions for scenarios sharing the same history. An example of such a scenario tree is shown in Figure~\ref{fig:scenario_tree_diagram}.
% \vspace{-1em}

% \vspace{-.5em}
\subsubsection{Diffusion Scenario Tree}
\label{sec:diffusion_ST}
% \vspace{-.5em}
The \emph{Diffusion Scenario Tree} method, described in Algorithm~\ref{alg:diffusion_tree}, constructs the tree, stage-by-stage, from the root using TimeGrad. For stage \(t\), at each node, TimeGrad samples \(M\) forecasts \(\{\hat{o}^{s,(i)}_t\}_{i=1}^M \subseteq \mathbb{R}^{H \times D}\) for the stochastic component, from the conditional distribution of future timesteps, conditioned on past observations from the node’s predecessors, over the stage horizon \(H\). These forecasts are clustered using K-means into \(K\) groups, with centroids serving as representative predictions for the nodes. Branch probabilities are computed as \(p_k = \frac{1}{M} \sum_{i=1}^M \mathds{1}(\ell_i = k)\), and the top-\(L\) child nodes are retained based on the highest cumulative probabilities of the paths from the root to these nodes at stage \(t\). The retained nodes' probabilities are normalized to sum to 1. Each cluster centroid is used as the input for forecasting at the next stage. Therefore \emph{DST} recursively samples future trajectories and generates nodes based only on tree paths, ensuring the non-anticipativity constraint (decisions depend only on observed history) by construction.

A drawback of our method is the computationally expensive diffusion inference, as Algorithm~\ref{alg:diffusion_tree} requires sampling $M$ trajectories at every node. For the application considered (and many applications in the field of operations research, where actions have significant impact), this is acceptable, since decisions are made on a daily sampling interval. For real-time applications, alternative diffusion variants (e.g., DDIM~\cite{song2020denoising}, DPM-Solver++~\cite{lu2025dpm}) would be more appropriate and are worth investigating in future work.
% \vspace{-.5em}
\subsubsection{Solving the Multistage Stochastic MPC}
% \vspace{-.5em}
Given a scenario tree \(\mathcal{T}\), we solve the multistage stochastic MPC to optimize decisions across \( N \) stages:
\vspace{-.5em}

\begin{equation}
	\begin{aligned}
	\underset{\{a^i_t\}}{\text{maximize}} & \quad \sum_{t=k_0}^{k_0+N-1} \sum_{i \in \mathcal{T}_t} \pi_i R(o^{d,i}_t, o^{s,i}_t, a^i_t) \\
	\text{subj. to} \quad & o^{d,1}_{k_0} = o^d_{k_0}, \quad o^{s,1}_{k_0} = o^s_{k_0}, \\
	& \begin{cases}
		o^{d,i}_{t+1} = f(o^{d,\mathrm{pre}(i)}_t, a^{\mathrm{pre}(i)}_t), \\
		o^{s,i}_{t+1} = o^{s,\mathrm{pre}(i)}_{t+1},
	\end{cases} \; \forall i \in \mathcal{T}_{t+1} \setminus \{1\}, \\
	& a^i_t = a^j_t \; \forall i, j \in \mathcal{T}_t \text{ for same history}, a^i_t \in \mathcal{A} \; \forall i, t.
\end{aligned}
\end{equation}

% \vspace{-.5em}
Here, \(\mathcal{T}_t\) denotes nodes at stage \(t\), with probability \(\pi_i\), observation \((o^{d,i}_t, o^{s,i}_t)\), and decision \(a^i_t\). The deterministic component evolves via \(f\), while the stochastic part is forecasted.

% \vspace{-.5em}
\section{Results}
\label{sec:results}
% \vspace{-1em}
We evaluate the \emph{Diffusion Scenario Tree (DST)} framework on the energy arbitrage task in New York's day-ahead electricity market, where a Battery Energy Storage System (BESS) optimizes energy transactions based on forecasted prices. The environment is modeled as a POMDP, with observations consisting of battery state-of-charge (SoC) and electricity prices, actions control battery charge, and rewards defined by the trading revenue minus degradation costs, defined thoroughly in~\cite{Zarifis2025}. In this problem, the deterministic component, battery State of Charge (SoC), evolves via known dynamics $f$, and the stochastic component's (energy prices) dynamics are estimated probabilistically by TimeGrad. Experiments use 10 months of hourly electricity price data, covering multiple seasons, to test on diverse market conditions\footnote{Information about the implementation of the models: \url{https://sites.google.com/view/stelios-zarifis-thesis-ntua/}, in Section "Experimental Results".}.

\begin{figure}[htbp]
	\centering
	\includesvg[width=.85\linewidth]{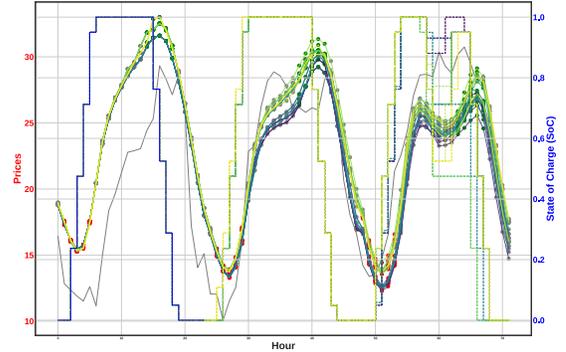}
	% \vspace{-.75em}
%	\fontsize{8.5}{10}\selectfont
	\caption{\textit{Strategy planned by Scenario tree-based MPC. Dotted lines depict the forecasted stochastic component, organized in a tree. Gray line shows the actual stochastic part evolution, blue line represents the optimized actions for the first stage, and dashed lines the optimized actions across branches.}}
	\label{fig:dst_mpc_example}
\end{figure}

Figure~\ref{fig:dst_mpc_example} shows the optimized actions by Stochastic MPC based on our \emph{DST} over a 3-day horizon. Dotted lines show forecasted prices in the scenario tree (branching daily), the gray line actual prices published after optimization, the blue line shows the applied optimal actions for the first stage, and dashed lines show how SoC evolves when applying the branch actions. \emph{DST}-based MPC applies the actions for the first stage and then re-plans, as new observations are available. Visually, we can see that the agent buys and sells energy at near-optimal points, generating profit from price differences across the horizon.
% \vspace{-1em}
% \vspace{-2.5em}

\subsection{\emph{DST}-based MPC vs. Standard MPC}
% \vspace{-.5em}
Table~\ref{tab:combined_dst_vs_others} compares \emph{DST}-guided MPC with benchmarks: Perfect MPC (full knowledge of the realized trajectories), Oracle MPC (perfect forecasts over the same finite horizon), and Monte Carlo SMPC (sampling multiple trajectories from the model’s distribution to estimate the expectation of the objective via Monte Carlo averaging, with its number of realized trajectories equal to \emph{DST}'s scenarios), the best-performing MPC using the diffusion-based forecaster from our previous work~\cite{Zarifis2025}, which also outperformed all implementations relying on conventional forecasters. \emph{DST} MPC achieves an average reward of $85.81$, which is comparable to and slightly better than MC SMPC (0.29\% improvement). This small margin should be interpreted in context: Monte Carlo SMPC is already a powerful model, achieving near-optimal performance (only ~2\% worse than Oracle MPC, which uses perfect forecasts), leaving limited room for improvement. Note that the considered problem is only mildly nonlinear; we expect DST to be more beneficial in capturing multimodal uncertainty in highly nonlinear systems, which we can explore in future work.
% \vspace{-1em}
\subsection{\emph{DST}-based MPC vs. LSTM- and VAR-based MPC}
% \vspace{-.5em}
Table~\ref{tab:combined_dst_vs_others} also shows \emph{DST} MPC versus LSTM and VAR Scenario Tree-based MPC. Our approach, surpasses both LSTM-based and VAR-based implementations by 18.0\% and 25.4\%, respectively. Consistently superior across all tests, \emph{DST}'s ability to discretize multimodal distributions provides a more accurate model to guide the MPC. Another point highlighting the benefit of using scenario tree-based MPC over standard MPC is that even the classical forecasters (LSTM, VAR), when used recursively in a scenario tree manner, provide more accurate actions, as multiple scenarios are considered instead of a single one in standard MPC. This leads to an improvement of approximately 91.2\% for the LSTM Scenario Tree-based MPC and 36.5\% for the VAR Scenario Tree-based MPC over the experiments of our previous work~\cite{Zarifis2025}.

\subsection{\emph{DST}-based MPC vs. Model-Free RL}
\emph{DST} significantly outperforms model-free Reinforcement Learning baselines (Deep Q-Networks), which achieve much lower rewards in this low-data, partial-model setting. Our best Deep Q-Network implementation (using the actual next 12 prices as state) achieved an average reward of only 26.92, compared to 85.81 for \emph{DST}-guided SMPC. The low RL performance stems from the low-data regime of our application (few thousand historical points) and partial observability (known battery dynamics but unknown future price evolution), which severely limits effective exploration and policy learning. RL training was also highly unstable, exhibiting large variance across multiple runs despite extensive tuning. In contrast, training our diffusion forecaster was extremely stable, with consistent predictive performance across all independent runs. This enables effective multi-step lookahead in \emph{DST}, whereas model-free RL struggles to extract useful policies from the same limited data. Related studies support that MPC often outperforms several RL algorithms when reliable (even partial) dynamics models are available~\cite{WANG2023120430,zhan2023comparing}.
% \vspace{-1em}

\begin{algorithm}[h!]
	\fontsize{8.5}{10}\selectfont
	\caption{Diffusion Scenario Tree}
	\label{alg:diffusion_tree}
	\KwIn{
		$\mathcal{T}$: empty tree; $\mathcal{O}$: observation history; $D$: tree depth; \\
		\fontsize{8}{10}\selectfont{\texttt{predictor}}\fontsize{8.5}{10}\selectfont: diffusion-based forecaster; $K$: clusters per stage; $M$: forecast samples; $H$: stage horizon; $L$: top-\(L\) children.
	}
	\KwOut{Scenario tree with forecasts \& probabilities}
	
	Init. queue \(Q \leftarrow \{(\text{root}, \mathcal{O}, 0)\}\), node\_counter = 0\;
	
	\While{$Q \neq \emptyset$}{
		\tcp{node, obs history, stage}
		$(node, \mathcal{O}, \ell) \leftarrow \text{Dequeue}(Q)$
		
		$node.\text{id}\!\leftarrow\!\text{node\_counter}$++$\!$ \tcp{Assign unique ID}
		
		\If{$\ell \ge D$}{\textbf{continue} \tcp{maximum depth reached}}
		
		\tcp{Sample conditioned to history}
		$\{\mathcal{F}^{(i)}\}_{i=1}^M = \texttt{predictor.sample}(\mathcal{O}^s, M)$
		
		\tcp{Cluster in groups}
		Reshape forecasts into $X \in \mathbb{R}^{M \times (H \cdot D)}$
		
		Cluster $X$ with K-means $\to$ labels $\{\ell_i\}$ 
		
		\For{$k = 1, \dots, K$}{
			\tcp{Representative forecast, branch \& cumulative prob}
			$\mu_k \leftarrow \text{centroid of cluster } k$ 
			
			$p_k \leftarrow \frac{1}{M} \sum_{i=1}^M \mathds{1}(\ell_i = k)$
			
			$P_k \leftarrow node.P \cdot p_k$
			
			Create child node $v_k$ with $(\mu_k, P_k)$\;
		}
		
		Sort $\{v_k\}$ by $P_k$, keep top $L$ \tcp*{Prune tree}
		
		Normalize probabilities: $P_k \leftarrow P_k / \sum_{j=1}^L P_j$
		
		\For{each kept $v_k$}{
			$\mathcal{O}' \leftarrow (\mathcal{O} \,||\, \mu_k)$ \tcp{Update obs history}
			
			Attach child $v_k$ to $node$
			
			Enqueue$(Q, (v_k, \mathcal{O}', \ell+1))$ \tcp{Expand}
		}
	}
	
	\Return{$\mathcal{T}$} \tcp{Final scenario tree}
\end{algorithm}
% \vspace{-.9em}
\section{Conclusion}
\label{sec:conclusion}
% \vspace{-1em}
We proposed the \emph{Diffusion Scenario Tree (DST)} framework, which constructs scenario trees using diffusion-based probabilistic forecasters like TimeGrad to improve multistage stochastic optimization. \emph{DST} recursively samples trajectories, to expand the scenario tree, and organizes them via clustering, ensuring non-anticipativity. The resulting structure not only enables better decision-making but also provides \emph{interpretability}, as each branch clearly represents a possible future scenario and the corresponding optimal actions. Evaluated on a real-world decision problem, our method outperforms the same scenario tree algorithms guided by more conventional predictive models, the standard MPC implementations guided by the same diffusion-based forecaster, and RL baselines. Future work includes problem-driven scenario tree generation, so that the generated scenarios align with the problem's structural characteristics and constraints~\cite{Fairbrother2019,Fairbrother2023}.

\begin{table}[!htbp]
	\centering
	\caption{Performance comparison (higher is better). DST-guided SMPC is comparable to and slightly better than Monte Carlo SMPC using the same diffusion forecaster (D-I MPC), while the scenario-tree versions of VAR and LSTM substantially outperform their standard point-forecast counterparts. Bold values indicate the better performer within each forecaster pair.}
	\label{tab:combined_dst_vs_others}
	\resizebox{\columnwidth}{!}{%
		\begin{tabular}{l|cc|cc|cc|cc}
			\toprule
			\textbf{Month} & \textbf{Perfect} & \textbf{Oracle} & \textbf{D-I} & \textbf{DST} & \textbf{LSTM} & \textbf{LSTM ST} & \textbf{VAR} & \textbf{VAR ST} \\
			 & \textbf{MPC} & \textbf{MPC} & \textbf{MPC} & \textbf{MPC} & \textbf{MPC} & \textbf{MPC} & \textbf{MPC} & \textbf{MPC} \\
			\midrule
			2018-06 & $99.98$ & $97.13$ & $90.05$ & $\mathbf{94.47}$ & $74.84$ & $\mathbf{87.94}$ & $\mathbf{65.84}$ & $60.51$ \\
			2018-07 & $147.46$ & $141.56$ & $134.87$ & $\mathbf{140.00}$ & $38.13$ & $\mathbf{116.23}$ & $68.81$ & $\mathbf{107.92}$ \\
			2018-08 & $123.11$ & $115.31$ & $111.28$ & $\mathbf{112.63}$ & $3.59$ & $\mathbf{109.21}$ & $61.02$ & $\mathbf{97.11}$ \\
			2018-09 & $108.44$ & $102.32$ & $99.67$ & $\mathbf{100.20}$ & $32.56$ & $\mathbf{96.09}$ & $60.77$ & $\mathbf{79.63}$ \\
			2018-10 & $104.05$ & $95.36$ & $\mathbf{102.13}$ & $101.17$ & $79.75$ & $\mathbf{81.43}$ & $60.85$ & $\mathbf{70.20}$ \\
			2019-04 & $60.35$ & $54.62$ & $\mathbf{52.19}$ & $45.00$ & $\mathbf{50.13}$ & $36.14$ & $30.77$ & $\mathbf{34.25}$ \\
			2020-12 & $68.85$ & $59.31$ & $57.41$ & $\mathbf{59.92}$ & $\mathbf{54.41}$ & $54.37$ & $30.65$ & $\mathbf{55.97}$ \\
			2021-01 & $64.03$ & $59.00$ & $55.87$ & $\mathbf{58.27}$ & $42.87$ & $\mathbf{52.56}$ & $37.63$ & $\mathbf{51.84}$ \\
			2021-02 & $109.34$ & $108.33$ & $\mathbf{98.20}$ & $95.71$ & $-46.61$ & $\mathbf{50.99}$ & $56.99$ & $\mathbf{78.03}$ \\
			2021-03 & $56.94$ & $56.35$ & $\mathbf{53.95}$ & $50.72$ & $\mathbf{50.88}$ & $42.37$ & $28.20$ & $\mathbf{49.12}$ \\
			\midrule
			\textbf{Sum} & $942.57$ & $889.30$ & $855.63$ & $\mathbf{858.09}$ & $380.54$ & $\mathbf{727.32}$ & $501.54$ & $\mathbf{684.58}$ \\
			\textbf{Average} & $94.26$ & $88.93$ & $85.56$ & $\mathbf{85.81}$ & $38.05$ & $\mathbf{72.73}$ & $50.15$ & $\mathbf{68.46}$ \\
			\bottomrule
		\end{tabular}
	}
\end{table}

%\begin{table}[!htbp]
%	\centering
%	\caption{\fontsize{9pt}{11pt}\selectfont \emph{DST} SMPC vs MFRL in 10 months (higher is better).}
%	\label{tab:dst_vs_rl}
%	% \vspace{-1em}
%	\resizebox{0.73\linewidth}{!}{%
%		\fontsize{7}{10}\selectfont
%		\begin{tabular}{l|cc}
%			\toprule
%			\textbf{Model} & \textbf{Sum} & \textbf{Average} \\
%			\midrule
%			\textbf{DQN} \tiny ($32 \times 64 \times 32$) & $55.76$ & $5.58$ \\
%			\textbf{DQN} \tiny ($64 \times 128 \times 64$) & $\underline{269.15}$ & $\underline{26.92}$ \\
%			\textbf{DQN} \tiny ($128 \times 256 \times 128$) & $151.80$ & $15.18$ \\
%			\textbf{DQN} \tiny ($64 \times 128 \times 128 \times 64$) & $51.02$ & $5.10$ \\
%			\textbf{DQN} \tiny ($64 \times 128 \times 256 \times 128 \times 64$) & $196.27$ & $19.63$ \\
%			\midrule
%			\midrule
%			\textbf{DST SMPC} & $\mathbf{858.09}$ & $\mathbf{85.81}$ \\
%			\bottomrule
%		\end{tabular}
%	}
%\end{table}

\bibliographystyle{IEEEbib}
\bibliography{refs}

\vspace*{\fill}

\begin{minipage}{\columnwidth} 
	\noindent % Ensures the first line of the combined content is not indented
	
	% Left minipage for the text (e.g., 70% of the column width)
	\begin{minipage}{0.75\linewidth} % [t] aligns top of minipage; \linewidth here refers to the parent minipage's width (which is \columnwidth)
		\raggedright % Text in this minipage is left-aligned (standard for multi-line text)
		{\section*{Acknowledgment} \footnotesize This project is funded by the European Union under Horizon Europe (grant No. 101136568 - HERON).}
	\end{minipage}%
	% Small horizontal space between the text minipage and the image minipage
	%		\hspace{0.5em}%/
	% Right minipage for the image (e.g., 25% of the column width)
	\begin{minipage}{0.25\linewidth} % [t] aligns top of minipage; \linewidth here refers to the parent minipage's width
		%			\raggedleft % Align the image to the right within its own minipage
		\includegraphics[width=2.06cm]{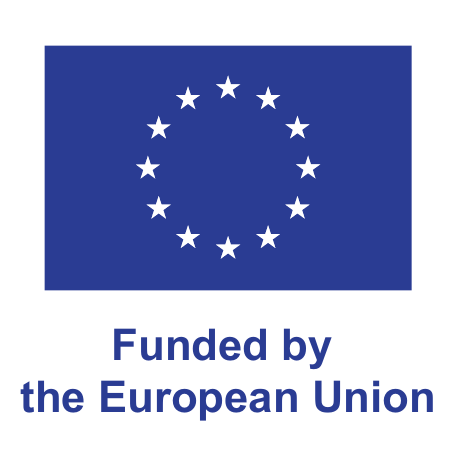}
	\end{minipage}
\end{minipage}

\end{document}